\documentclass[11pt]{article}

\usepackage{acl}

\usepackage{times}
\usepackage{latexsym}

\usepackage[T1]{fontenc}
\usepackage[utf8]{inputenc}

\usepackage{microtype}

\usepackage{inconsolata}

\usepackage{graphicx}
\usepackage{booktabs,multirow,siunitx,makecell}
\usepackage[inkscapelatex=false]{svg}
\usepackage{times}
\usepackage{latexsym}
\usepackage{amsmath}
\usepackage{amsfonts}
\usepackage{amssymb}
\usepackage{stmaryrd}
\usepackage{algorithm}
\usepackage{algpseudocode}
\usepackage[utf8]{inputenc}
\usepackage{enumitem}
\usepackage{array}
\usepackage{booktabs}
\usepackage{microtype}
\usepackage{multirow}
\usepackage{inconsolata}
\usepackage{graphicx}
\usepackage{comment}
\usepackage{xcolor}
\usepackage{tikz}
\usetikzlibrary{positioning, shapes, arrows.meta, fit}
\title{Entropy-Gated Branching for Efficient Test-Time Reasoning}
\author{
    Xianzhi Li\textsuperscript{}, 
    Ethan Callanan\textsuperscript{},
    Abdellah Ghassel\textsuperscript{},
    Xiaodan Zhu\textsuperscript{}
\\
    \textsuperscript{}Department of Electrical and Computer Engineering \& Ingenuity Labs Research Institute \\ Queen's University\\
    \{li.xianzhi, e.callanan, abdellah.ghassel, xiaodan.zhu\}@queensu.ca
}

\begin{document}
\maketitle
\begin{abstract}
    Test-time compute methods can significantly improve the reasoning capabilities and problem-solving accuracy of large language models (LLMs). However, these approaches require substantially more computational resources, with most compute wasted on exploring low-diversity branches where the model already exhibits high confidence. We observe that a small subset of uncertain reasoning steps has a disproportionately large impact on final prediction accuracy, and branching at these critical junctures tends to yield more diverse and higher-quality candidate reasoning steps. We propose Entropy-Gated Branching (EGB), which branches only at high-uncertainty steps and prunes expansions with a lightweight verifier. On mathematical and financial reasoning benchmarks, EGB improves accuracy by 22.6\% over standard inference while operating 31\%-75\% faster across math benchmarks than test-time beam search with higher performance. Our results show that dynamic resource allocation during inference can substantially improve both efficiency and effectiveness, offering a more scalable pathway to enhanced LLM reasoning capabilities. We release our code and tools here\footnote{\url{https://github.com/JXL884/entropy_gated_branching}}
\end{abstract}

\section{Introduction}

\begin{figure*}[h!]
    \centering
    \includegraphics[width=1\linewidth]{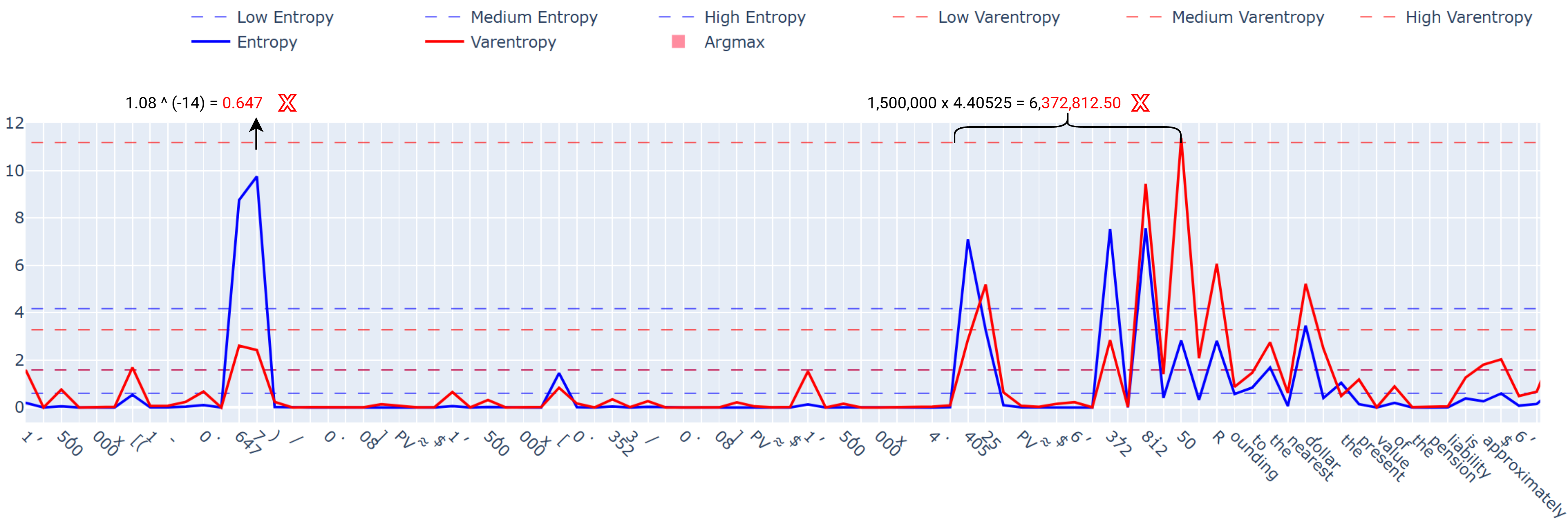}
    \caption{Entropy \textit{(blue)} and Varentropy \textit{(red)} distribution of \texttt{Llama-3.2-1B-instruct} solving a Chartered Financial Analyst (CFA) problem, demonstrating uncertainty spikes aligning with the model's mistakes.}
    \label{fig:main1}
\end{figure*}

Language models have demonstrated remarkable capabilities across a diverse range of tasks \cite{bang:23, li2023chatgpt, mahfouz2024state}, yet persist in making fundamental errors due to hallucinations or flawed reasoning \cite{huang2024survey, tong2024can, Farquhar2024Detecting, zhang2025llm, galitsky2025truth, Penny-Dimri2025Reducing}. These mistakes are especially detrimental in high-stakes domains such as finance. To address this issue, it is crucial to develop models that ensure outputs are both verifiable and interpretable, reducing the risk of critical mistakes.

Consequently, understanding the underlying causes of these errors and devising practical mitigation strategies is vital for building more reliable language models. Many studies \cite{xie2025logic, yu2025dapo, wei2025swe} have demonstrated that fine-tuning language models using rewards or high-quality data can significantly enhance their performance. However, these approaches require substantial time and computational resources to retrain the model. Recently, the research community has shifted focus to test-time compute (TTC) as a complementary approach to improving model performance without modifying trained weights \cite{bi2024forest, liang2024improving, hfblog, snell2025scaling}. TTC methods offer the compelling advantage of enhancing model capabilities without modifying trained weights, enabling performance scaling at inference time. Many TTC methods operationalize extra computation through extensive resampling \cite{wang2022self}, or searching multiple reasoning branches at every step and selecting the most promising continuation \cite{xie2023self, snell2025scaling}.

However, test-time search algorithms could take minutes to explore all potential reasoning chains before generating the final answer. This is not user-friendly and poses a significant barrier to real-world deployment. To understand when that extra test-time computation is worthwhile, we analyzed next-token logit entropy on mathematical reasoning benchmarks \cite{wang20258020rulehighentropyminority}. As shown in Figure \ref{fig:main1}, models are considerably more error-prone during entropy spikes. Since a spike implies a broad posterior over tokens, allocating extra expansion precisely at these high-entropy steps yields maximal diversity and a better chance to rescue an otherwise faulty trajectory; conversely, when entropy is low, the model exhibits high confidence, making further expansion less fruitful since it would likely produce similar continuations. Empirically, we found that 39 out of 50 occurrences that coincide with a high-entropy spike lead to a flawed step of reasoning. These observations suggest that high entropy moments can indicate effective points to branch the model's reasoning, rather than wasting computation by branching at every step. This raises the question: Is it possible to improve TTC efficiency by selectively expanding beams while still maintaining the performance benefits of beam search?

We propose Entropy-Gated Branching (EGB), a novel TTC inference method that uses the entropy of the model’s predicted logit distribution to identify optimal branching points, instead of expanding at every step. By adopting entropy as an uncertainty signal and combining it with external feedback models to jointly steer the search, EGB concentrates the budget on high-impact uncertainty moments, while confident beams continue with standard sampling, avoiding unnecessary computational overhead. EGB demonstrates the ability to outperform other test-time search algorithms while achieving substantial reductions in inference time and computational costs.

    % At the core of our approach is combining entropy as a gating mechanism together with external feedback models to jointly steer the branching process to \textit{branch only when necessary.} 
\section{Related Work}

\paragraph{Uncertainty Quantification}
Handling uncertainty in language models is crucial for ensuring reliable decision-making and response generation. Prior work has explored sampling multiple responses to detect inconsistencies as a measure of uncertainty \cite{manakul2023selfcheckgpt, abbasi2024believe, wagner2024black}. However, despite their effectiveness, these methods cannot assess the reliability of a single response and require multiple iterations, making them inefficient for real-world applications. An alternative approach leverages token-probability methods \cite{fadeeva2024fact, duan2024shifting}, which assess uncertainty at the token level without requiring multiple resampling steps. Our work builds on token-level uncertainty methods by using entropy as a real-time signal to guide branching decisions during generation, rather than as a post-hoc uncertainty estimate. This allows EGB to dynamically allocate computation based on the model's confidence at each step.

\paragraph{Diversity and Efficiency in Decoding}
A fundamental challenge in generation is balancing diversity with computational efficiency. Standard beam search maintains multiple hypotheses but explores uniformly at each step, often generating redundant or low-quality alternatives \cite{holtzman2019curious, kulikov2018importance, wang2022self, snell2025scaling}. Diverse beam search methods \cite{vijayakumar2018diverse} address this by explicitly encouraging dissimilarity among beams. Self-evaluation-guided stochastic beam search (SEGBS) \cite{xie2023self} introduced self-evaluation guidance via stochastic beam search, but still expands at every timestep regardless of model confidence. In contrast, EGB fundamentally differs from prior work by making branching itself uncertainty-aware rather than treating all generation steps equally.

\paragraph{Test-Time Computation}
Recent research has demonstrated that enhancing computational resources during inference can yield significant performance gains without retraining models \cite{snell2025scaling, liu2025can, ji2025test}. Most test-time computing methods rely on generating multiple solutions through repeated sampling and selecting the best answer. Search algorithms like Monte-Carlo Tree Search \cite{zhou2023language, zhang2023planning, liu2024don, koh2024tree}, guided beam search \cite{xie2024self}, and hybrid search \cite{snell2025scaling} show that inference-time search can improve performance across various tasks. Another approach \cite{muennighoff2025s1} appends "wait" tokens to control computational budget during reasoning. However, these methods typically apply uniform computational effort across all generation steps. EGB addresses this inefficiency by identifying when and where additional computation is most beneficial through entropy-based gating, achieving the performance benefits of test-time search with substantially reduced computational overhead.

\paragraph{Process Reward Models and Feedback}
In language models, external feedback mechanisms are essential for guiding generation toward higher-quality outputs \cite{pan2024feedback, li2025survey, estevez2025evaluation}. Recent work has explored aligning models with human preferences \cite{AIoa2400196, scheurer2023training}, prompt optimization \cite{zhou2022large, liu2024large, ma2024large}, and error refinement \cite{jimichi2023feedback, kirstein2024s}. Process reward models (PRMs) have emerged as particularly effective for step-wise reasoning tasks, evaluating intermediate steps rather than only final answers \cite{lightman2023let, uesato2022solving}. EGB integrates PRMs as a critical component for ranking candidate branches generated at uncertainty spikes, ensuring that expanded beams are selected based on reasoning quality rather than arbitrary heuristics.

% \begin{figure*}[h!]
%     \centering
%     \input{latex/diagrams/method-diagram.tex}
%     \caption{Illustration of the branching mechanism}
%     \label{fig:method}
% \end{figure*}

\begin{figure*}[h!]
    \centering
    \includegraphics[width=0.9\linewidth]{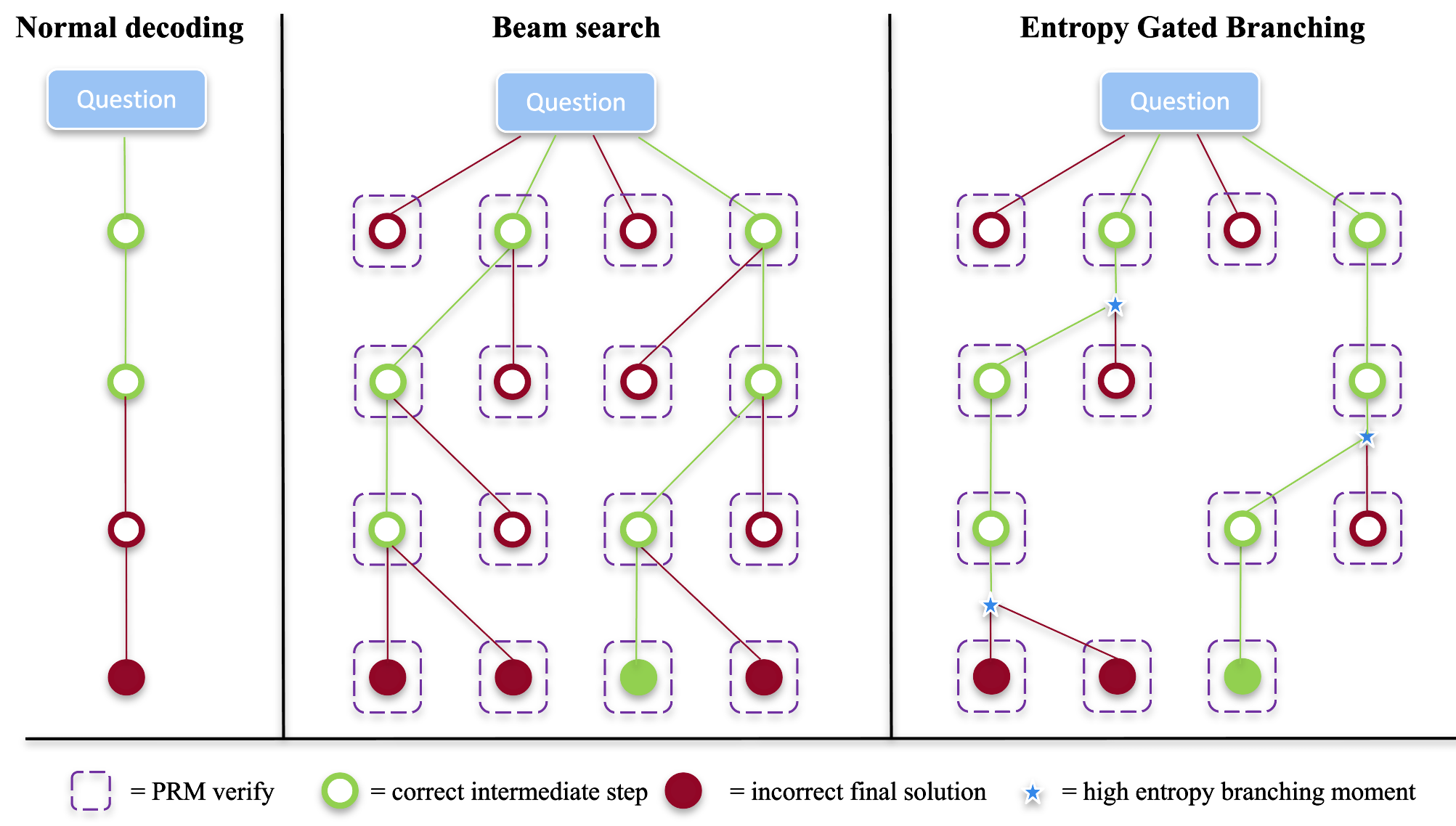}
    \caption{Illustration of Entropy-Gated Branching; \textbf{Left:} normal decoding where the model flows naturally. \\ \textbf{Middle:} traditional beam search samples \(KW\) candidates and uses PRM scores keep the top \(K\) beams. \\ \textbf{Right:} EGB expands uncertain beams at high entropy moments and generates confident beams normally.}
    \label{fig:main2}
\end{figure*}

\section{Methodology}
We study step-wise search at test time for sequence-of-thought generation. A solution is represented as a sequence of reasoning steps
\(y_{1:T}=(s_1,\ldots,s_T)\), where each step \(s_t\) is a semantically coherent chunk (e.g., delimited by ``.\verb|\n|'' or ``\verb|\n\n|''). At step \(t\), standard beam search maintains \(K\) active beams; for each beam it proposes \(W\) candidate continuations for step \(t{+}1\), scores all \(K\times W\) candidates with a process reward model (PRM), and keeps the top \(K\). This uniform policy expands \(W\) branches at every step and for every beam, regardless of how confident the model is about the next tokens. In practice, many expansions are near-duplicates when the next-token distribution is sharp, wasting compute without adding useful diversity.

Entropy-Gated Branching (EGB) replaces this uniform policy with selective expansion. Instead of branching everywhere, EGB monitors uncertainty during decoding and only allocates extra branches at positions where the model is uncertain. Section~\ref{subsec:egb} details the gating and rollback mechanism; Section~\ref{subsec:formalexp} formalizes the resulting candidate pool and budget.

\subsection{Entropy-Gated Branching}\label{subsec:egb}
While decoding the next reasoning step for a beam, we compute the entropy of a model's logit distribution at timestep $t$ as:

\begin{equation}
H_t = -\sum_{i=1}^{V} p_{i,t} \log_2 p_{i,t},
\end{equation}
where $p_{i,t}$ denotes the predicted probability of token $i$ at step $t$, and $V$ is the vocabulary size. High entropy indicates a flatter distribution and thus higher uncertainty.

The decision to branch is controlled by a crucial hyperparameter: the entropy threshold, $\tau$. This threshold is not universal; its optimal value is domain-specific and depends on both the base model's calibration and the nature of the task. For instance, highly creative or open-ended tasks might naturally exhibit higher entropy, whereas logical deduction tasks may have a lower baseline entropy. Consequently, $\tau$ must be tuned on a held-out validation set to find the right balance between computational efficiency (favoring a higher $\tau$ to branch less) and exploration (favoring a lower $\tau$ to capture more uncertainties). We detail our tuning process and analyze the sensitivity of our results to $\tau$ in Section~\ref{sec:results}.

\paragraph{Certain Beams \(\mathcal{C}_t = \{ b \mid H_t^{(b)} \le \tau \}\)}
When the entropy for a given beam $b$ is below the threshold $\tau$, it signals that the model is confident in its next-token prediction. In this case, EGB avoids wasteful exploration by not creating unnecessary branches for high-confidence parts of the solution. We use the default sampling configuration to generate the most likely next reasoning chunk.

\paragraph{Uncertain Beams \(\mathcal{U}_t = \{ b \mid H_t^{(b)} > \tau \}\)}
When entropy exceeds the threshold, it indicates a critical decision point where the model is uncertain and multiple reasoning paths may be viable. These are the moments where exploration is most valuable. We employ a rollback-and-branch strategy to precisely locate and expand at uncertainty points. When generating a continuation sequence for beam $i$, we monitor the entropy at each token position. If at any position $t' > t$ the entropy $H_{t'}^{(i)}$ exceeds $\tau$ during generation, we identify the earliest such position:
\begin{equation}
t^* = \min\{t' \mid t' > t \text{ and } H_{t'}^{(i)} > \tau\}
\end{equation}
We then roll back the generation to position $t^*$ and create $W$ diverse candidate branches starting precisely from this high-entropy token. Formally, the rollback operation truncates the generated sequence:
\begin{equation}
s^{(i)} \leftarrow s^{(i)}_{1:t^*-1}
\end{equation}
and initiates branching at position $t^*$, yielding candidate set $\{c_{t^*,1}^{(i)}, \dots, c_{t^*,W}^{(i)}\}$. This rollback mechanism ensures that branching occurs exactly at the first moment of uncertainty rather than at arbitrarily chosen checkpoints, concentrating search budget precisely where the model's confidence wavers.

\subsection{Formalizing Selective Expansion} \label{subsec:formalexp}
This process yields a combined pool of candidates, \(\mathcal{P}_t\), from all active beams. The composition of this pool is determined by whether a beam belongs to the uncertain set \(\mathcal{U}_t\) or the certain set \(\mathcal{C}_t\):
\begin{equation} \label{eq:selective_expansion}
    \mathcal{P}_t = \bigcup_{i=1}^{K}
    \begin{cases}
        \{c_{t,1}^{(i)}, \dots, c_{t,W}^{(i)}\}, & \text{if } i \in \mathcal{U}_t \\
        \{c_{t,1}^{(i)}\}, & \text{if } i \in \mathcal{C}_t
    \end{cases}
\end{equation}

% For uncertain beams requiring expansion, we employ temperature-scaled multinomial sampling to ensure diversity among the $W$ generated candidates. Specifically, for each uncertain beam $i \in \mathcal{U}t$, we sample from the probability distribution:
% \begin{equation}
% p'{j,t} = \frac{\exp(l_{j,t}/\tau_{temp})}{\sum_{k=1}^{V} \exp(l_{k,t}/\tau_{temp})}
% \end{equation}
% where $l_{j,t}$ represents the raw logit for token $j$ at timestep $t$, and $\tau_{temp}$ is a temperature parameter that controls the diversity of sampling. A lower temperature encourages more focused sampling around high-probability tokens, while higher temperatures promote greater exploration of the token space.

The total number of candidates generated at step \(t\) is therefore at most \(|\mathcal{P}_t| = K + (W-1)|\mathcal{U}_t|\). Since \(|\mathcal{U}_t| \le K\), this is always less than or equal to the \(KW\) candidates produced by standard beam search. The computational complexity of selective expansion scales as $O(|\mathcal{U}_t| \cdot W \cdot V)$ for candidate generation, compared to $O(K \cdot W \cdot V)$ for uniform beam search. In practice, since $|\mathcal{U}_t|$ is much smaller than $K$ due to the selective nature of branching, EGB achieves substantial computational savings while maintaining search effectiveness.

\subsection{Beam Evaluation}
Before proceeding to quality assessment, EGB implements a duplicate detection mechanism to avoid wasting computational resources on scoring identical continuations. Since both uncertain beams and certain beams contribute to the candidate pool $\mathcal{P}_t$, identical or near-identical sequences can emerge from different sources, particularly when the model exhibits strong preferences for specific token sequences. Therefore, we employ exact string matching at the token level and remove duplicate branches first, and retain the remaining diverse branches for further evaluation.
After collecting candidate continuations, assessing their quality is critical, as it directly influences the overall performance of the base model. A strong feedback model ensures that the generation follows a coherent and logical trajectory while minimizing errors.
We employ a Process Reward Model that evaluates the quality of reasoning steps rather than just final outcomes. The PRM assigns scores to partial solutions based on their logical consistency, mathematical accuracy, and adherence to sound reasoning principles. For a candidate continuation $c_{t,j}^{(i)}$, the PRM score is computed as:
\begin{equation}
s_{PRM}(c_{t,j}^{(i)}) = f_{PRM}(\text{context}, c_{t,j}^{(i)})
\end{equation}
where $f_{PRM}$ represents the trained process reward model that takes the problem context and the candidate continuation as input. Finally, we sort the scored candidates in descending order and select the top-$K$ highest-scoring beams as the active beams for continued branching in the next step.
\section{Experiments} \label{exp}

\paragraph{Datasets}
We evaluate on mathematical reasoning benchmarks of varying difficulty, including AIME \cite{aopsAIME}, MATH-500 \cite{hendrycks2021measuringmathematicalproblemsolving}, and GSM8K \cite{cobbe2021trainingverifierssolvemath}. In addition to general math benchmarks, we also include financial math datasets to test our method in a domain-specific setting. We include financial math from CFA mock exams purchased from AnalystPrep \cite{analystprep}, covering two program levels. CFA exams assess rigorous financial and quantitative reasoning in high-stakes scenarios where errors can cause significant losses. We present more details about the dataset in the Appendix \ref{sec:appendix}.

\paragraph{Language Models}
We implement our method on two families of models, \texttt{Qwen3} \cite{Yang2025Qwen3} and \texttt{Llama3} \cite{touvron2023llamaopenefficientfoundation}, at 3 sizes each. For \texttt{Qwen3} we use \texttt{Qwen3-1.7B}, \texttt{Qwen3-4B}, and \texttt{Qwen3-8B}, all in "non-thinking" mode \cite{Yang2025Qwen3}. For \texttt{Llama3} we use \texttt{Llama-3.2-1B-instruct}, \texttt{Llama-3.2-3B} \texttt{-instruct}, and \texttt{Llama-3.1-8B-instruct}. Unless otherwise noted, we adopt the recommended sampler settings. In addition, we explore \emph{thinking-mode} decoding for \texttt{Qwen3} in Appendix~\ref{app:thinkingmode}. This setting produces substantially longer reasoning traces and different efficiency--accuracy trade-offs.

\paragraph{Feedback Models}
To provide reliable feedback, we use \texttt{Qwen2.5-Math-PRM-7B}, which is the most popular PRM model on HuggingFace up to date. This model is specifically trained to assess reasoning quality by assigning a numerical score that reflects the correctness of the inference. Unlike the instruct model, which relies on a qualitative analysis to provide a branch selection, this model directly provides a probabilistic evaluation, where a designated probability serves as a quantitative score for ranking the branches. Higher scores indicate stronger logical consistency and correctness.  To assess robustness to verifier choice, we additionally evaluate EGB with an alternative PRM, \texttt{RLHFlow/Llama3.1-8B-PRM-Deepseek-Data}, in Appendix~\ref{app:prm_sensitivity}.

\paragraph{Baselines}
We benchmark EGB against several strong test-time reasoning methods. Self-consistency \cite{wang2022self} generates multiple independent reasoning paths and aggregates them via majority voting. Self-evaluation-guided stochastic beam search \cite{xie2023self} incorporates model self-evaluation into the beam search process. Step-wise beam search with external PRM feedback \cite{snell2025scaling} employs deterministic decoding to retain the top-k candidates at each reasoning step.

\paragraph{Branching Settings}
For results reported in Table~\ref{tab:main}, we use a beam size $K$ of 4, a beam width $W$ of 4 for all datasets, and a sample budget of 16 for Self-consistency. Detailed analyses of the threshold values and branch numbers can be found in Section~\ref{sec:ablation}.

\paragraph{Infrastructure}
All experiments were performed on H100 GPUs with 80GB VRAM. Each experiment was run on one GPU, utilizing 20-70 GB of GPU memory, depending on the model size and sequence length.

\begin{table*}[ht!]
\centering
\setlength{\tabcolsep}{6pt}
\renewcommand{\arraystretch}{0.95}
\resizebox{\textwidth}{!}{%
\begin{tabular}{ccccccc}
\toprule
\multirow{2}{*}{\textbf{Model}} & \multirow{2}{*}{\textbf{Method}} & \multicolumn{2}{c}{\textbf{CFA}} & \multirow{2}{*}{\textbf{AIME}} & \multirow{2}{*}{\textbf{MATH-500}} & \multirow{2}{*}{\textbf{GSM8K}} \\
\cmidrule(lr){3-4}
& & \textbf{Level I} & \textbf{Level II} & & & \\
\midrule\midrule
\multirow{5}{*}{\texttt{Qwen3-1.7B}}
 & Standard         & 48.78\% & 39.77\% & 6.67\%  & 70.40\% & 76.80\% \\
 & Self-consistency & 56.89\% & 44.89\% & \textbf{13.33\%} & 74.23\% & 84.23\% \\
 & SEGBS             & 55.94\% & 45.45\% & 7.78\%  & 69.60\% & 80.67\% \\
 & Beam Search      & 58.72\% & 50.57\% & 10.00\% & 73.80\% & 86.58\% \\
 & EGB              & \textbf{61.50\%} & \textbf{51.70\%} & 11.11\% & \textbf{77.40\%} & \textbf{89.46\%} \\
\midrule
\multirow{5}{*}{\texttt{Qwen3-4B}}
 & Standard         & 67.28\% & 54.55\% & 17.78\% & 82.80\% & 90.90\% \\
 & Self-consistency & 66.67\% & 56.36\% & 17.78\% & 78.60\% & 92.42\% \\
 & SEGBS             & 71.61\% & 57.39\% & 17.78\% & 80.20\% & 93.18\% \\
 & Beam Search      & \textbf{72.69\%} & 52.27\% & \textbf{21.11\%} & 81.20\% & 93.18\% \\
 & EGB              & 69.72\% & \textbf{60.80\%} & 18.60\% & \textbf{83.80\%} & \textbf{94.01\%} \\
\midrule
\multirow{5}{*}{\texttt{Qwen3-8B}}
 & Standard         & 73.50\% & 51.70\% & 18.89\% & 82.40\% & 92.72\% \\
 & Self-consistency & 76.56\% & 59.66\% & 18.89\% & 78.40\% & 92.72\% \\
 & SEGBS             & 76.61\% & 58.52\% & 17.78\% & 82.60\% & 93.10\% \\
 & Beam Search      & 76.00\% & 57.39\% & 18.89\% & 82.40\% & 94.54\% \\
 & EGB              & \textbf{77.73\%} & \textbf{61.36\%} & \textbf{22.22\%} & \textbf{84.20\%} & \textbf{95.45\%} \\
\midrule\midrule
\multirow{5}{*}{\texttt{Llama-3.2-1B-instruct}}
 & Standard         & 30.44\% & 25.57\% & 2.22\%  & 21.80\% & 36.39\% \\
 & Self-consistency & 37.56\% & 31.82\% & 3.33\%  & 42.80\% & 52.39\% \\
 & SEGBS             & 38.17\% & 31.82\% & \textbf{11.11\%} & 42.58\% & 58.21\% \\
 & Beam Search      & \textbf{43.83\%} & 31.82\% & 3.33\%  & \textbf{45.80\%} & \textbf{63.00\%} \\
 & EGB              & 42.28\% & \textbf{34.09\%} & 3.33\%  & 44.20\% & 61.92\% \\
\midrule
\multirow{5}{*}{\texttt{Llama-3.2-3B-instruct}}
 & Standard         & 43.17\% & 36.36\% & 5.56\%  & 39.00\% & 72.33\% \\
 & Self-consistency & 52.17\% & 40.89\% & \textbf{11.11\%} & 56.39\% & 72.30\% \\
 & SEGBS             & 46.61\% & 41.48\% & \textbf{11.11\%} & 50.84\% & 85.24\% \\
 & Beam Search      & 51.11\% & 38.64\% & 10.00\% & 55.80\% & \textbf{87.49\%} \\
 & EGB              & \textbf{52.28\%} & \textbf{42.61\%} & \textbf{11.11\%} & \textbf{56.80\%} & 84.69\% \\
\midrule
\multirow{5}{*}{\texttt{Llama-3.1-8B-instruct}}
 & Standard         & 49.17\% & 38.07\% & 4.44\%  & 41.80\% & 83.09\% \\
 & Self-consistency & 53.83\% & 50.89\% & 10.00\% & 57.84\% & 90.45\% \\
 & SEGBS             & 54.39\% & 48.18\% & \textbf{11.11\%} & 56.20\% & 90.65\% \\
 & Beam Search      & \textbf{56.94\%} & 43.75\% & \textbf{11.11\%} & 60.20\% & \textbf{91.36\%} \\
 & EGB              & 55.78\% & \textbf{51.14\%} & \textbf{11.11\%} & \textbf{62.20\%} & 90.67\% \\
\bottomrule
\end{tabular}}

\caption{Performance of EGB compared with beam search, self-evaluation-guided stochastic beam search (SEGBS), self-consistency, and standard decoding on finance and math benchmarks.}
\label{tab:main}
\end{table*}

\section{Results} \label{sec:results}
Our main results are presented in Table~\ref{tab:main}, outlining the performance of each model across the five benchmark datasets. EGB demonstrates substantial improvements over standard base inference, achieving an average gain of 18.4\% across all models and benchmarks with absolute accuracy gains of 3-8 percentage points (pp) in most settings. The method consistently delivers competitive or superior performance compared to beam search while achieving these results with significantly reduced computational cost and time, as detailed in Section~\ref{sec:time}. Our analysis reveals interesting patterns across different model families, scales, and reasoning domains.

\subsection{Main results}

\paragraph{General Math vs. Financial Reasoning} EGB consistently outperforms standard decoding on all domains, but there are nuanced patterns between models and domains. For smaller models (1-2B parameters), EGB yields comparable improvements across both domains, suggesting that when base model capabilities are limited, both financial and mathematical reasoning benefit equally from increased exploration. However, as model scale increases, financial reasoning tasks maintain robust gains from EGB while general mathematical problems show diminishing returns. At the largest scales tested (8B parameters), financial tasks benefit at approximately twice the rate of general mathematical problems.

This scale-dependent pattern likely reflects fundamental differences in how uncertainty manifests across these reasoning domains. Financial reasoning problems typically require integrative processing that combines numerical computation, contextual interpretation, and domain-specific procedural knowledge. In such scenarios, EGB's exploration strategy can effectively navigate the complex interaction between different reasoning components.

\paragraph{Model Families} The Qwen models establish strong baseline performance, with even the smallest \texttt{Qwen3-1.7B} model substantially outperforming comparable Llama models across all benchmarks. For instance, \texttt{Qwen3-1.7B} achieves 48.78\% on CFA Level I compared to \texttt{Llama-3.2-1B}'s 30.44\%, and 70.40\% on MATH-500 versus 21.80\% for Llama. This performance advantage persists consistently across all model scales. However, the Llama family demonstrates exceptional responsiveness to EGB, achieving an average improvement of 31.2\% over standard decoding compared to Qwen's more modest 12.8\% gains. The Llama models show particularly dramatic improvements on certain tasks. For example, \texttt{Llama-3.2-1B-instruct} improves by +8.52pp on CFA Level II, while the stronger baseline \texttt{Qwen3-1.7B} gains +11.93pp on the same task. Remarkably, EGB's substantial gains enable Llama models to bridge much of the performance gap with their Qwen counterparts, often matching or approaching Qwen's performance despite starting from significantly weaker foundations.

\paragraph{Model Size Trends} EGB's advantages over beam search increase monotonically with model size. Small models (1-1.7B) show substantial relative gains from EGB (+26.1\% over baseline) but exhibit mixed results against beam search (-0.8pp average). Medium models (3-4B) achieve balanced performance with EGB winning 60\% of comparisons against beam search (+3.2\% relative advantage). Large models (8B) demonstrate EGB's strongest performance profile, with a 70\% win rate and substantial improvements over beam search (+5.8\% relative advantage).

This scaling pattern suggests that larger models do not reduce exploration headroom but rather enhance the quality and effectiveness of exploration itself. The superior performance of larger models with EGB indicates that increased reasoning capability amplifies the benefits of strategic exploration, as these models can both generate more sophisticated alternative approaches and more accurately evaluate their relative merits. While small models benefit from any additional reasoning paths, larger models can more effectively generate diverse, high-quality alternatives and better distinguish between promising and unpromising directions.

\paragraph{Baseline Comparisons} EGB demonstrates consistent advantages over existing decoding strategies. Compared to self-consistency, EGB achieves superior or comparable performance on 83\% of benchmark-model combinations. Leveraging Qwen family models, EGB consistently outperforms self-consistency by an average of 3.2pp, with particularly notable gains on MATH-500 (+3.17pp for \texttt{Qwen3-1.7B}) and GSM8K (+5.23pp for \texttt{Qwen3-1.7B}). Against SEGBS, EGB shows clear advantages on harder benchmarks, winning 70\% of comparisons on CFA Level II. This demonstrates the value of leveraging external feedback from process reward models over model self-evaluation, particularly for smaller models that may struggle to reliably assess their own reasoning quality. The performance gap becomes more pronounced on complex tasks where accurate self-feedback requires sophisticated capabilities that smaller models often lack. The comparison with beam search reveals EGB's efficiency advantage: while achieving comparable or superior accuracy on 67\% of tasks, EGB does so with substantially lower computational overhead. EGB's superior performance stems from its selective expansion strategy. Unlike beam search, which expands at every time step and risks selecting low-quality or similar beams through PRM that lead to incorrect trajectories, EGB strategically targets high-entropy regions to diversify reasoning paths more effectively. These improvements demonstrate that EGB's entropy-guided exploration effectively captures reasoning uncertainty in ways that complementary approaches cannot fully replicate.

\begin{figure*}[ht!]
    \centering
    \includegraphics[width=1\linewidth]{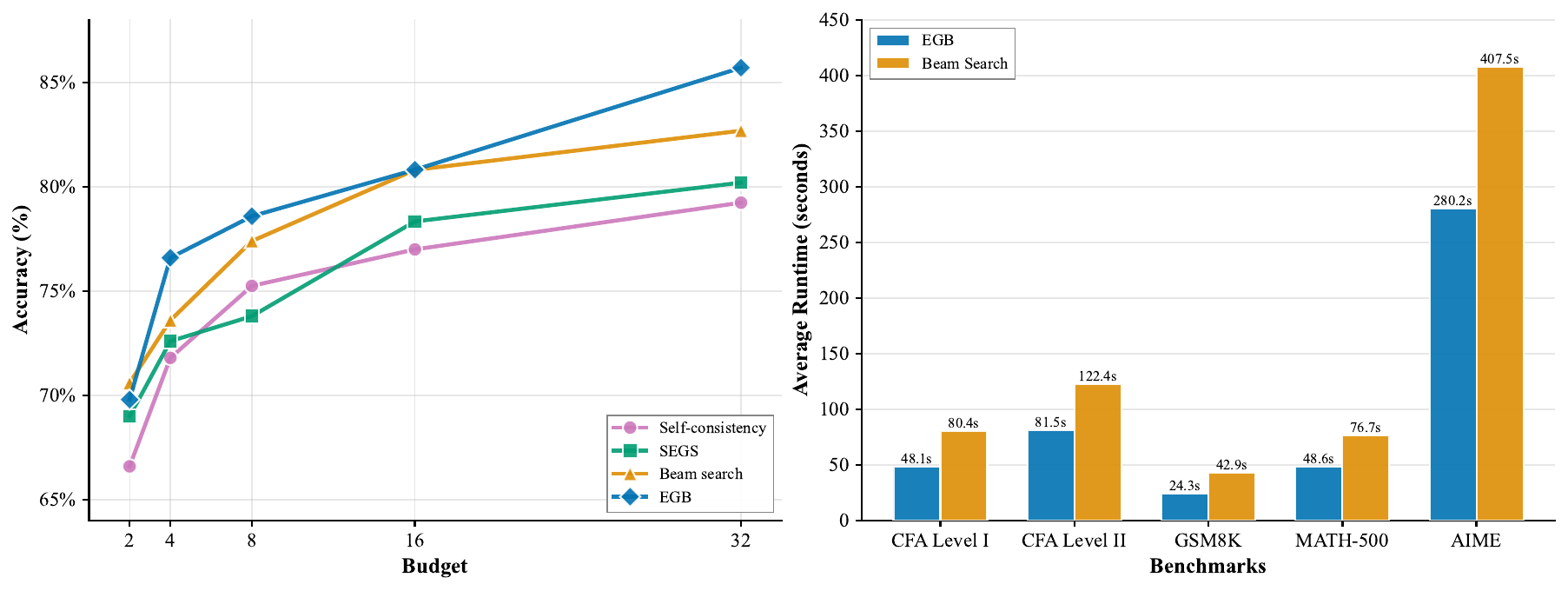}
    \caption{(Left) Budget–accuracy scaling across inference methods. \emph{Budget:} for Self-Consistency, the budget is the total number of sampled solutions; for beam-style methods (SEGBS, Beam Search, and EGB), the budget is the product of beam width and the number of expansions. (Right) Average runtime per benchmark for EGB and Beam Search, as it proves to be the most competitive baseline.}

    \label{fig:budget}
\end{figure*}

\paragraph{Computational Budget Analysis}
To assess the efficiency and scalability of different decoding methods, we evaluate performance across varying computational budgets ranging from 2 to 32. As illustrated in Figure~\ref{fig:budget}, EGB demonstrates superior scaling characteristics compared to alternative approaches. At low budgets (2-4), EGB achieves competitive performance with beam search and SEGBS, trailing beam search by only 0.8pp at budget=2 but surpassing it by 3.0pp at budget=4. This early crossover point highlights EGB's efficient utilization of limited computational resources. The advantage becomes more pronounced at higher budgets: at budget=32, EGB reaches 85.71\% accuracy, outperforming beam search by 3.02pp, SEGBS by 5.51pp, and self-consistency by 6.47pp. Notably, while all methods show performance saturation at higher budgets, EGB maintains a steeper improvement trajectory, suggesting more effective allocation of additional computation. Self-consistency exhibits the most gradual scaling curve, gaining only 12.64pp from budget=2 to budget=32, compared to EGB's 15.91pp improvement over the same range. This analysis demonstrates that EGB not only achieves superior final performance but does so with better computational efficiency, making it particularly valuable in resource-constrained settings where maximizing accuracy per unit of computation is critical.

\subsection{Computational Efficiency Analysis}\label{sec:time}

\paragraph{Run-Time Efficiency} Beyond accuracy improvements, EGB delivers substantial computational efficiency advantages over beam search, achieving 31-75\% speedup across all evaluated benchmarks, as shown in Figure~\ref{fig:budget}, while maintaining superior or competitive performance. Because Self-Consistency and SEGBS consistently underperform EGB across budgets and benchmarks, we report runtime comparisons only against Beam Search, the strongest alternative. This efficiency gain represents a significant practical advantage that enhances EGB's value proposition for real-world deployment scenarios. Unlike traditional accuracy-efficiency trade-offs, EGB achieves both improved performance and reduced computational cost simultaneously. For financial reasoning tasks, EGB delivers 33-40\% speedups alongside 2-4pp accuracy improvements. Even on challenging mathematical benchmarks where accuracy gains are modest, the substantial time savings provide clear value for time-sensitive deployments.

\paragraph{GPU Memory Efficiency} EGB provides significant memory efficiency advantages by avoiding the storage overhead associated with maintaining large numbers of parallel candidate sequences. Since most generation steps involve primarily confident predictions, EGB consistently operates with significantly fewer active candidates than traditional beam search, directly translating to proportional memory savings for intermediate state storage, attention caches, and token embeddings. The memory savings compound over longer sequences, where the cumulative effect of selective branching becomes increasingly pronounced.

\section{Conclusion}
We introduced Entropy-Gated Branching (EGB), a selective expansion strategy that dynamically allocates computational resources based on model uncertainty during generation. We addressed the limitation in current test-time compute methods by recognizing that not all generation steps require equal computational investment. By using entropy as a gating mechanism to trigger branching only at high-uncertainty decision points, EGB achieves a combination of improved accuracy and reduced computational cost. Our comprehensive evaluation across mathematical reasoning benchmarks demonstrates that EGB consistently delivers substantial speedup while maintaining superior performance over other test-time search algorithms. The method proves particularly effective on moderately challenging tasks where models possess the requisite knowledge but struggle with reliable reasoning paths, achieving substantial accuracy improvements on financial and mathematical reasoning problems. By focusing compute on where it matters the most, EGB offers a pragmatic path to robust, cost-aware test-time reasoning at scale.

\section*{Limitations}\label{sec:limitations}
While EGB demonstrates consistent improvements across mathematical and financial reasoning benchmarks, a few considerations remain. First, although our threshold sensitivity analysis reveals broad "safe zones" where performance is stable, automated threshold selection mechanisms could further enhance ease of deployment. Second, our evaluation focuses on domains where process reward models are well-established; extending EGB to domains with less mature verifiers (e.g., open-ended generation or commonsense reasoning) remains future work. Finally, combining EGB with thinking-mode decoding incurs higher computational costs due to longer reasoning traces, requiring practitioners to balance accuracy gains against latency constraints in resource-sensitive applications.

\section*{Acknowledgement}

This research was funded in part by the Faculty Research Awards of J.P. Morgan AI Research. The authors are solely responsible for the contents of the paper and the opinions expressed in this publication do not reflect those of the funding agencies. We also thank Mathieu Sibue, Antony Papadimitriou, Zhiqiang Ma, and Xiaomo Liu for helpful discussions and valuable feedback that contributed to this work.

\bibliography{custom}

\newpage
\appendix

\section{Appendix}
\label{sec:appendix}

\subsection{Algorithm}
\begin{figure*}[!htb]
    \centering
    \includegraphics[width=0.8\linewidth]{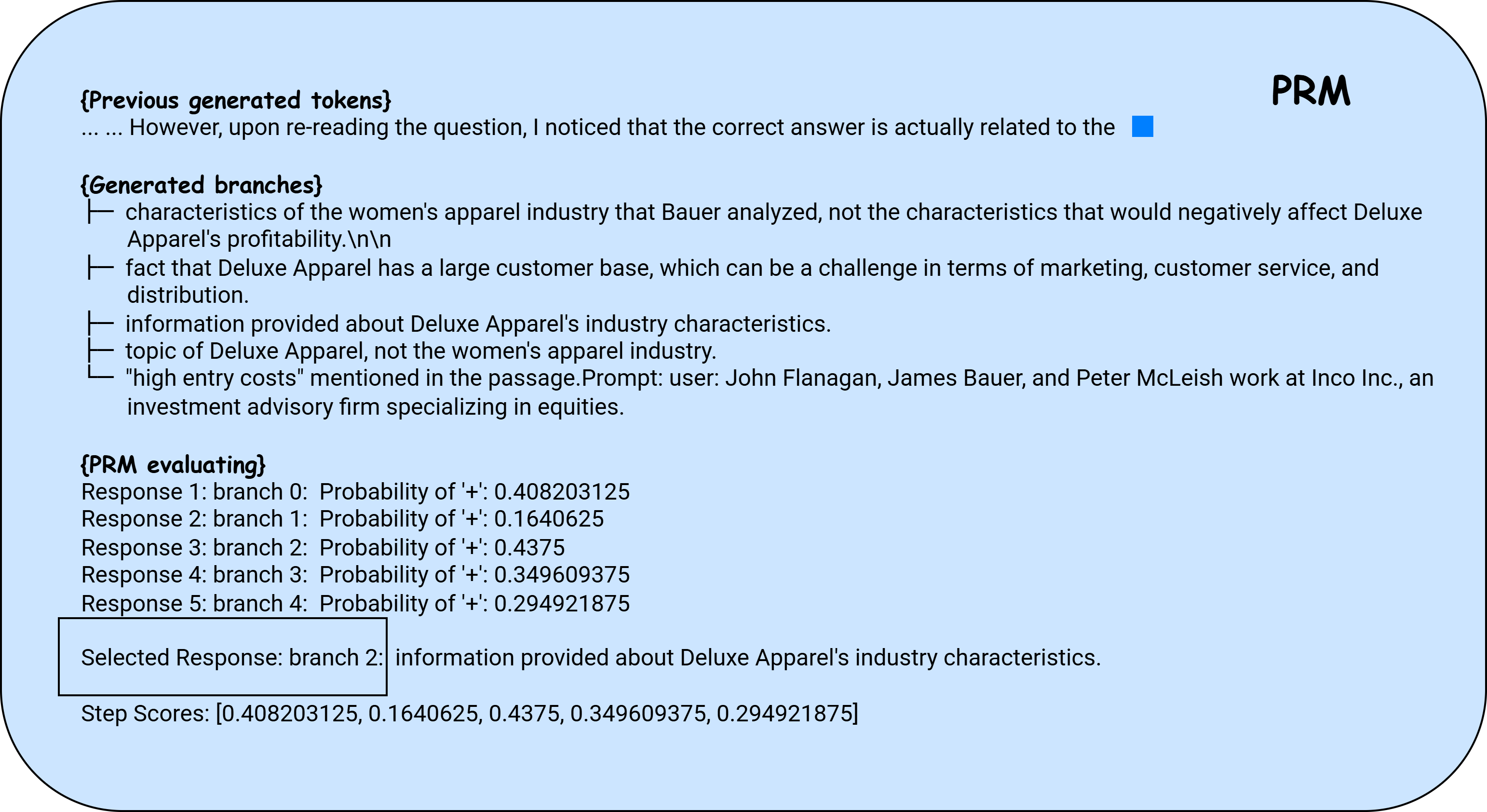}
    \caption{One example of PRM response}
    \label{fig:PRM_appendix}
\end{figure*}

We present the EGB algorithm, which shows how our method dynamically allocates computational resources by branching only when the model exhibits high uncertainty, leading to more efficient test-time computation compared to traditional beam search. We also show an example figure of how the  PRM model selects the top-performing beams in Figure~\ref{fig:PRM_appendix}.

\begin{algorithm}[h!]
\caption{Entropy-Gated Branching (EGB)}
\label{alg:egb}
\begin{algorithmic}[1]
\Require Model $M$, Problem $x$, Beam size $K$, Beam width $W$, Entropy threshold $\tau$, Process Reward Model $f_{PRM}$, Maximum steps $T$
\Ensure Best scoring solution

\State Initialize beam set $\mathcal{B}_0 = \{x\}$ with single beam containing problem $x$
\State $t \leftarrow 1$

\While{$t \leq T$ and beams not terminated}
    \State $\mathcal{P}_t \leftarrow \emptyset$ \Comment{Initialize candidate pool}
    
    \For{each beam $b \in \mathcal{B}_{t-1}$}
        \State Compute entropy: $H_t^{(b)} = -\sum_{i=1}^{V} p_{i,t}^{(b)} \log_2 p_{i,t}^{(b)}$
        
        \If{$H_t^{(b)} \leq \tau$} \Comment{Certain beam}
            \State Generate single candidate: $c_{t,1}^{(b)} \sim M(\cdot | b)$
            \State $\mathcal{P}_t \leftarrow \mathcal{P}_t \cup \{c_{t,1}^{(b)}\}$
        \Else \Comment{Uncertain beam}
            \For{$j = 1$ to $W$}
                \State Generate candidate: $c_{t,j}^{(b)} \sim M(\cdot | b, \tau_{temp})$ \Comment{With temperature scaling}
                \State $\mathcal{P}_t \leftarrow \mathcal{P}_t \cup \{c_{t,j}^{(b)}\}$
            \EndFor
        \EndIf
    \EndFor
    
    \State Remove duplicates from $\mathcal{P}_t$ using string matching
    
    \For{each candidate $c \in \mathcal{P}_t$}
        \State Compute PRM score: $s_{PRM}(c) = f_{PRM}(\text{context}, c)$
    \EndFor
    
    \State Sort candidates by PRM scores in descending order
    \State $\mathcal{B}_t \leftarrow$ top-$K$ candidates from $\mathcal{P}_t$
    \State $t \leftarrow t + 1$
\EndWhile

\State \textbf{return} highest scoring beam from $\mathcal{B}_t$
\end{algorithmic}
\end{algorithm}

\paragraph{Dataset Details}
We evaluate our methods on competition and general mathematical reasoning benchmarks of varying difficulty. First, we include an AIME evaluation set drawn from the AI-MO dataset \cite{aopsAIME}. This corpus contains all 90 problems from the 2022–2024 American Invitational Mathematics Examination (AIME I \& II each year) \cite{aopsAIME}. Each AIME test is widely viewed as hard in difficulty between grade-school word problems and full Olympiad-style proofs \cite{maaAIME,wikiAIME}.

We also assess performance on the MATH benchmark \cite{hendrycks2021measuringmathematicalproblemsolving}, which consists of competition-level problems (AMC, AIME, and other sources) spanning a wide range of high-school mathematics topics and difficulty levels; for all experiments, we use 500 held-out test questions. Finally, we report results on GSM8K \cite{cobbe2021trainingverifierssolvemath}, a widely used grade-school math word-problem benchmark designed to probe multi-step arithmetic reasoning in language models, utilizing 1,319 test questions in our experiments.

In addition to general math benchmarks, we also include financial math datasets to test our method in a domain-specific setting. CFA exams are renowned for their rigorous assessment of financial, quantitative, and analytical reasoning, making them an excellent proxy for evaluating model performance in high-stakes, real-world financial decision-making scenarios. Mistakes in the financial questions can lead to significant losses, reinforcing the need for precise and reliable reasoning. As official CFA exam questions are not publicly accessible, this study utilizes CFA mock exams purchased from AnalystPrep \cite{analystprep}, covering two levels of the CFA program. The dataset comprises multiple-choice and essay questions, each supplemented with corresponding answers, explanations, grading criteria, and metadata indicating the CFA topic associated with each question.

\begin{figure*}[th!]
    \centering
    \includegraphics[width=1.0\linewidth]{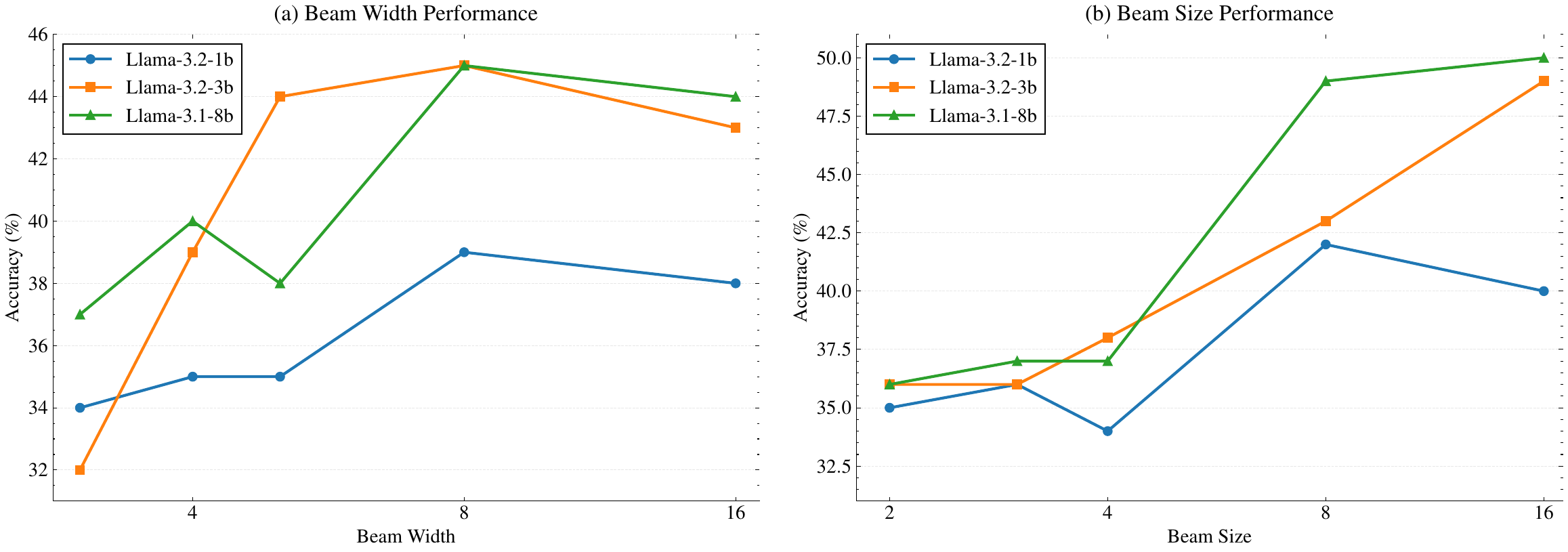}
    \caption{Impact of beam sizes \(K\) and beam widths \(W\) on model's performance, both numbers are scaled from 2 up to 16}
    \label{fig:beam-size-ablation}
\end{figure*}

\paragraph{LLM-based Evaluation}
Final answer correctness was determined using a powerful external language model, Gemini 2.5 Flash, as an impartial judge \cite{geminiteam2024gemini}. This approach circumvents the limitations of rigid evaluation methods like exact string matching or regex-based parsers, which are often brittle and can penalize semantically correct answers due to minor stylistic or formatting variations. For each problem, the LLM judge was provided with the original question, the model's complete generated response, and the ground-truth solution, and was then prompted to assess whether the final answer was correct. This method ensures a robust and fair assessment by focusing on the semantic correctness of the reasoning outcome rather than superficial syntactic differences, providing a more reliable measure of the model's true problem-solving capabilities.

\subsection{Ablation Studies}\label{sec:ablation}

\paragraph{Beam Size \(K\)}\label{sec:efficiency} We analyze beam size from \(K\)=2 to \(K\)=16, revealing that larger beam sizes consistently improve performance, with the most substantial gains occurring between \(K\)=2 and \(K\)=8 across all model scales. The improvement is particularly pronounced for larger models, with \texttt{Llama-3.1-8B} showing dramatic gains from 36.2\% to 48.8\% accuracy as beam size increases. This pattern confirms that maintaining multiple active reasoning branches helps preserve solution diversity and prevents premature convergence to suboptimal paths. However, the performance plateaus at \(K\)=16, suggesting diminishing returns. This is likely due to a trade-off in PRM-based ranking: as the PRM may favour clusters of similar partial solutions, additional beams may not contribute new information, limiting the benefits of further expansion.

\paragraph{Beam Width \(W\)} The beam width parameter, controlling how many new candidates are generated at each branching point, shows more modest but consistent improvements as W increases from 2 to 16. While the effect is less dramatic than beam size, larger models still benefit meaningfully from increased exploration breadth, with performance gains of 1-2 percentage points across the range. The relatively smaller impact of beam width compared to beam size suggests that maintaining diverse active branches is more critical than generating many candidates at each branching point. This indicates that EGB's effectiveness depends more on preserving multiple reasoning trajectories over time than on exhaustive local exploration at individual decision points.

\begin{table*}[ht!]
\centering
\small
\setlength{\tabcolsep}{5pt}
\begin{tabular}{ll lccccc}
\toprule
Base Model & PRM & Method & CFA-L1 & CFA-L2 & MATH-500 & GSM8K & Avg \\
\midrule
\multirow{4}{*}{\texttt{Llama-1B}}
& \multirow{2}{*}{\texttt{Llama3.1-8B-PRM}} & Beam Search & 41.06 & 32.95 & \textbf{39.60} & 57.39 & 42.75 \\
& & EGB    & \textbf{42.22} & \textbf{33.66} & 39.40 & \textbf{62.40} & \textbf{44.42} \\
\cmidrule(lr){2-8}
& \multirow{2}{*}{\texttt{Qwen2.5-Math-PRM}} & Beam Search & \textbf{43.83} & 31.82 & \textbf{45.80} & \textbf{63.00} & \textbf{46.11} \\
& & EGB    & 42.28 & \textbf{34.09} & 44.20 & 61.92 & 45.62 \\
\midrule
\multirow{4}{*}{\texttt{Qwen3-1.7B}}
& \multirow{2}{*}{\texttt{Llama3.1-8B-PRM}} & Beam Search & 58.91 & 47.16 & 71.19 & 86.03 & 65.82 \\
& & EGB    & \textbf{60.80} & \textbf{50.98} & \textbf{75.06} & \textbf{87.81} & \textbf{68.66} \\
\cmidrule(lr){2-8}
& \multirow{2}{*}{\texttt{Qwen2.5-Math-PRM}} & Beam Search & 58.72 & 50.57 & 73.80 & 86.58 & 67.42 \\
& & EGB    & \textbf{61.50} & \textbf{51.70} & \textbf{77.40} & \textbf{89.46} & \textbf{70.02} \\
\bottomrule
\end{tabular}
\caption{Sensitivity to PRM choice. Accuracy of Beam Search and EGB under two process reward models: \texttt{Qwen2.5-Math-PRM-7B} (default in main experiments) and an alternative verifier \texttt{RLHFlow/Llama3.1-8B-PRM-Deepseek-Data} (shown as \texttt{Llama3.1-8B-PRM}). \textbf{Bold} marks the better method within each base-model/PRM setting for each benchmark.}
\label{tab:prm_sensitivity}
\end{table*}

\paragraph{PRM Quality Sensitivity and Domain Generality}
\label{app:prm_sensitivity}

We use \texttt{Qwen2.5-Math-PRM-7B} as the process reward model for candidate scoring and ranking throughout our main experiments. In this study, we assess the robustness of EGB to the choice of PRM by replacing it with an alternative verifier, \texttt{RLHFlow/Llama3.1-8B-PRM-Deepseek-Data}, which is trained on different data and may exhibit different calibration properties. We evaluate two base models (\texttt{Llama-1B} and \texttt{Qwen3-1.7B}) on four benchmarks (CFA-L1, CFA-L2, MATH-500, GSM8K), while keeping the same decoding/branching configuration as in the main results (beam size $K=4$, beam width $W=4$). In all settings, the PRM is used only to score and rank candidates, consistent with Algorithm \ref{alg:egb}.

Table~\ref{tab:prm_sensitivity} reports accuracy for Beam Search with PRM feedback and EGB under both PRMs. We observe that swapping the PRM can affect absolute performance, particularly on math benchmarks like MATH-500, GSM8K, indicating that verifier quality/calibration impacts downstream selection. Importantly, EGB remains effective under both PRMs. For \texttt{Qwen3-1.7B}, the gain of EGB over Beam Search is consistent across PRMs (average +2.84pp with \texttt{Llama3.1-8B-PRM} and +2.60pp with \texttt{Qwen2.5-Math-PRM} across the four benchmarks). For \texttt{Llama-1B}, EGB improves over Beam Search with \texttt{Llama3.1-8B-PRM} (average +1.67pp), and is comparable under \texttt{Qwen2.5-Math-PRM} (average -0.49pp), with improvements on CFA-L2 but decreases on some math benchmarks. Overall, these results suggest that PRM choice primarily affects absolute accuracy, while entropy-gated branching provides a largely complementary benefit to verifier strength.

\begin{table*}[ht!]
\centering
\small
\setlength{\tabcolsep}{5pt}
\begin{tabular}{llccccccccc}
\toprule
Model & Task &
$\tau{=}0$ & $0.5$ & $1.0$ & $1.5$ & $2.0$ & $2.5$ & $3.0$ & $4.0$ & $\tau{=}\infty$ \\
\midrule
\multirow{2}{*}{\texttt{Qwen3-1.7B}}
& CFA-L2    & 50.57 & 49.43 & 47.16 & \textbf{52.27} & 50.00 & 48.30 & 44.89 & 44.59 & 44.89 \\
& MATH-500  & 73.80 & 64.33 & \textbf{74.94} & \textbf{75.00} & \textbf{74.80} & \textbf{77.20} & \textbf{77.60} & \textbf{76.40} & 74.23 \\
\midrule
\multirow{2}{*}{\texttt{Qwen3-4B}}
& CFA-L2    & 52.27 & \textbf{56.25} & \textbf{56.82} & \textbf{56.25} & 55.68 & \textbf{57.95} & \textbf{56.82} & 56.09 & 56.16 \\
& MATH-500  & 81.20 & 73.06 & 80.97 & \textbf{82.58} & \textbf{83.88} & \textbf{84.51} & \textbf{83.40} & \textbf{81.89} & 78.60 \\
\midrule
\multirow{2}{*}{\texttt{Qwen3-8B}}
& CFA-L2    & 57.39 & 54.55 & 57.95 & \textbf{61.36} & \textbf{61.36} & 58.52 & 58.52 & \textbf{61.93} & 59.66 \\
& MATH-500  & 82.40 & 69.00 & 79.81 & \textbf{82.74} & \textbf{84.75} & \textbf{84.79} & \textbf{84.91} & \textbf{83.47} & 78.40 \\
\midrule
\multirow{2}{*}{\texttt{Llama-1B}}
& CFA-L2    & 31.82 & 27.84 & 26.70 & \textbf{33.52} & 31.82 & \textbf{32.95} & \textbf{35.80} & \textbf{36.36} & 31.82 \\
& MATH-500  & \textbf{45.80} & 42.60 & 42.20 & 40.00 & 39.40 & 37.58 & 38.81 & 36.50 & 42.80 \\
\midrule
\multirow{2}{*}{\texttt{Llama-3B}}
& CFA-L2    & 38.64 & \textbf{42.05} & \textbf{44.32} & 38.64 & 39.77 & 39.77 & \textbf{42.84} & \textbf{44.32} & 40.89 \\
& MATH-500  & 55.80 & 53.22 & 54.20 & 54.80 & 56.13 & \textbf{57.98} & \textbf{59.56} & 55.87 & 56.39 \\
\midrule
\multirow{2}{*}{\texttt{Llama-8B}}
& CFA-L2    & 43.75 & 44.32 & 42.05 & 47.16 & \textbf{52.50} & \textbf{55.11} & \textbf{51.70} & \textbf{53.98} & 50.89 \\
& MATH-500  & 60.20 & \textbf{62.91} & \textbf{61.18} & \textbf{61.41} & \textbf{61.76} & \textbf{66.67} & 59.26 & \textbf{63.03} & 57.84 \\
\bottomrule
\end{tabular}
\caption{Entropy-threshold sensitivity. Validation accuracy on finance (CFA-L2) and math (MATH-500) as we sweep $\tau \in \{0, 0.5, 1, 1.5, 2, 2.5, 3, 4, \infty\}$. Here $\tau{=}0$ branches at every step (equivalent to beam search) and $\tau{=}\infty$ disables branching (self-consistency). \textbf{Bold} indicates settings that outperform both endpoints ($\tau{=}0$ and $\tau{=}\infty$) for the same model and task.}
\label{tab:tau_sweep}
\end{table*}

\paragraph{Entropy-threshold sensitivity}
We study the sensitivity of EGB to the entropy threshold $\tau$ by sweeping $\tau \in \{0, 0.5, 1, 1.5, 2, 2.5, 3, 4, \infty\}$ for six base models on validation sets from two domains (finance: CFA Level II; math: MATH-500). This sweep induces a unified continuum that recovers standard baselines as special cases: $\tau{=}0$ branches at every step (equivalent to beam search), $\tau\in(0,\infty)$ performs selective branching only at high-entropy tokens (EGB), and $\tau{=}\infty$ disables branching entirely (equivalent to self-consistency / best-of-$n$ with voting). Across model families, we observe broad regions of $\tau$ that consistently outperform both endpoints, with accuracy typically varying by only 1--3 percentage points when $\tau$ is perturbed by $\pm 0.5$--$1.0$ around a reasonable choice, indicating a practical ``safe zone'' rather than a sharply tuned optimum. Moreover, the same $\tau$ ranges that perform well on MATH-500 also tend to perform well on CFA-L2 for a given model family, suggesting that $\tau$ behaves primarily as a model-family property rather than a task-specific hyperparameter. While our main configuration uses a single $\tau$ per model family for simplicity and ease of deployment, these sweeps also indicate that lightweight per-task adjustment of $\tau$ can yield additional gains (often on the order of 1--3 points) when tuning on a validation set is feasible.

\paragraph{Thinking-mode models} \label{app:thinkingmode}
We further evaluate EGB under \emph{thinking-mode} decoding on Qwen3 models, which produces substantially longer reasoning traces and is often used to elicit stronger reasoning performance. Table~\ref{tab:thinking_mode} compares thinking vs.\ non-thinking on CFA Level II. Enabling thinking mode significantly increases the base accuracy (e.g., \texttt{Qwen3-8B}: 51.70\% $\rightarrow$ 68.75\%) but also increases generation length and inference time (typically $2$--$5\times$ slower), which compounds the cost of test-time search. Under non-thinking decoding, EGB consistently improves over beam search (e.g., \texttt{Qwen3-8B}: 57.39\% $\rightarrow$ 61.36\%). Under thinking mode, EGB remains beneficial for the larger models (\texttt{Qwen3-4B}: 67.01\% $\rightarrow$ 72.16\%; \texttt{Qwen3-8B}: 70.38\% $\rightarrow$ 72.97\%), while the smallest model shows a smaller and less consistent gain (\texttt{Qwen3-1.7B}: 54.64\% $\rightarrow$ 54.00\%). Overall, thinking mode and entropy-gated branching are complementary in terms of accuracy, but they induce different efficiency--performance trade-offs: on \texttt{Qwen3-8B}, non-thinking+EGB achieves 61.36\% with $\sim$60s average runtime, whereas thinking+EGB achieves 72.97\% with $\sim$280s average runtime, reflecting the substantially higher cost of branching over long traces.

\begin{table*}[ht!]
\centering
\small
\setlength{\tabcolsep}{5pt}
\begin{tabular}{llccccc}
\toprule
Model & Mode & Base & Beam Search & EGB (ours) & $\Delta$EGB--BS & $\Delta$EGB--Base \\
\midrule
\multirow{2}{*}{\texttt{Qwen3-1.7B}}
& Non-thinking & 39.77 & 50.57 & \textbf{51.70} & +1.13 & +11.93 \\
& Thinking     & 53.98 & \textbf{54.64} & 54.00 & -0.64 & +0.02 \\
\midrule
\multirow{2}{*}{\texttt{Qwen3-4B}}
& Non-thinking & 54.55 & 58.27 & \textbf{60.80} & +2.53 & +6.25 \\
& Thinking     & 69.32 & 67.01 & \textbf{72.16} & +5.15 & +2.84 \\
\midrule
\multirow{2}{*}{\texttt{Qwen3-8B}}
& Non-thinking & 51.70 & 57.39 & \textbf{61.36} & +3.97 & +9.66 \\
& Thinking     & 68.75 & 70.38 & \textbf{72.97} & +2.59 & +4.22 \\
\bottomrule
\end{tabular}
\caption{CFA Level II accuracy (\%) for \texttt{Qwen3} models with thinking-mode enabled vs.\ disabled. We report base (greedy) decoding, Beam Search ($\tau{=}0$), and EGB (selective branching), along with absolute gains of EGB over Beam Search and over the base setting.}
\label{tab:thinking_mode}
\end{table*}

\begin{figure*}[htb]
    \centering
    \includegraphics[width=0.8\linewidth]{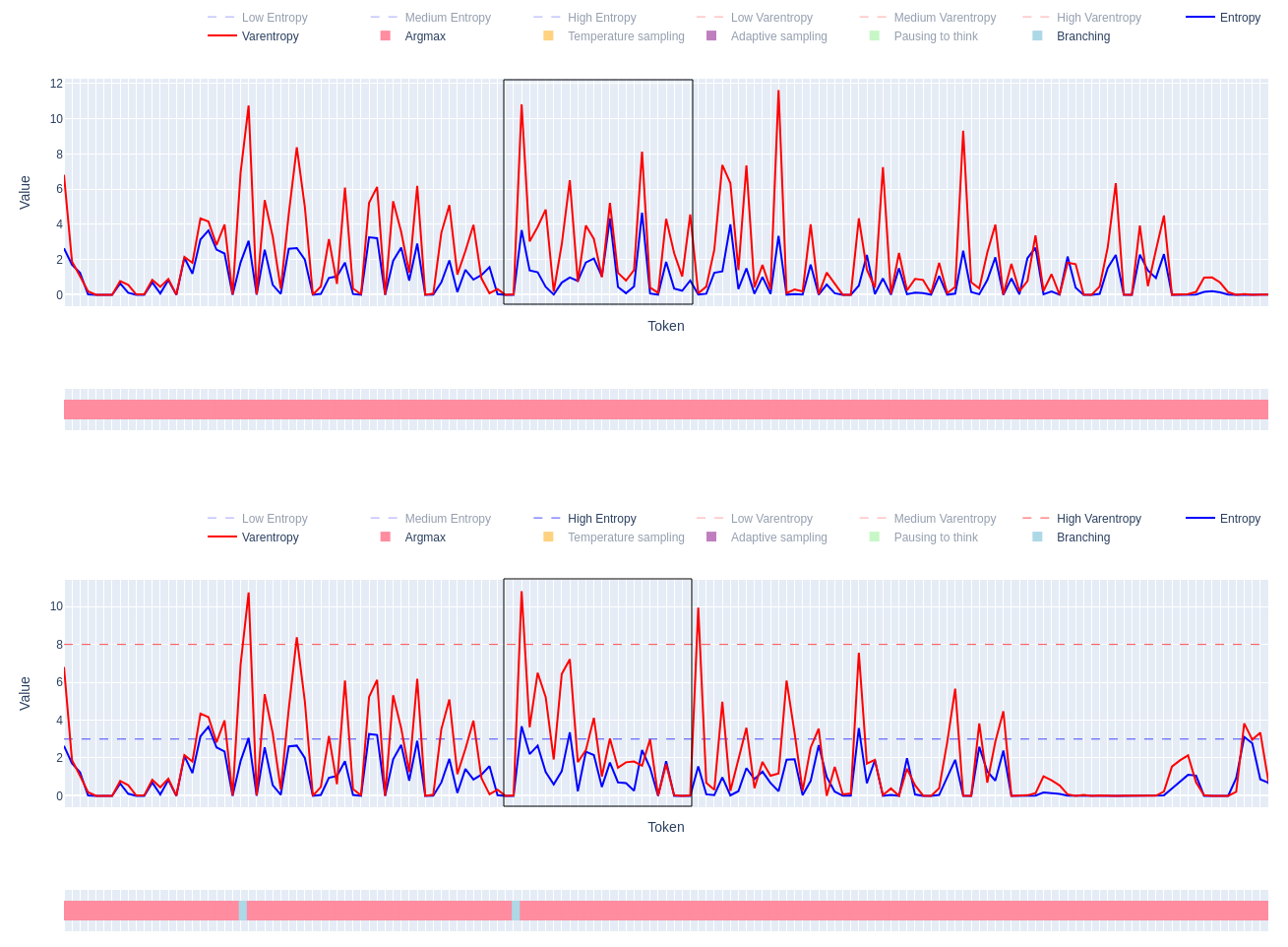}
    \caption{Entropy and varentropy plots along with sampler states of \texttt{Llama-3.1-8B-instruct} on a CFA II question using argmax sampling (top; incorrect answer) and entropy-aware branching (bottom; correct answer). Branching occurs at the start of the indicated region, resulting in a different reasoning path from argmax being selected. Notably, this causes a downtrend in both entropy and varentropy until the next spike, while the argmax plot remains higher and unstable. Branching points are shown as {\textcolor{blue}{\(\blacksquare\)}} in the red horizontal bar.}
    \label{fig:argmax_v_branching}
\end{figure*}

\subsection{Reduced Uncertainty} \label{app: C}
Figure \ref{fig:argmax_v_branching} presents the uncertainty distribution of \texttt{Llama-3.1-8B} on a CFA Level II question. The top plot shows the model’s behavior under argmax sampling, which ultimately yields an incorrect answer. In contrast, the bottom plot illustrates entropy-aware branching, resulting in a correct solution. Branching is triggered twice as shown in the horizontal bar at the indicated region. At the first branching point, our method select the same token as argmax, where the entropy and varentropy distribution are the same as argmax. However, at second branching point corresponds to a subsequent decline in both entropy and varentropy, suggesting that the model becomes more confident and stable as it continues reasoning with a different token selection. Meanwhile, in the argmax scenario (top plot), entropy and varentropy remain relatively high and fluctuate at the end of the generation, reflecting continuous uncertainty that ultimately leads to an erroneous conclusion.

\end{document}